\documentclass[0.5pt,letterpaper]{article}
\usepackage{fullpage}

\usepackage{microtype}
\usepackage{graphicx}
\usepackage{subfigure}
\usepackage{booktabs} % for professional tables

\usepackage{hyperref}
\usepackage{natbib}

\usepackage{graphics}

\usepackage{xfrac}
\usepackage{enumitem}
\usepackage{amsthm}
\usepackage{mathtools}
\usepackage[group-separator={,}]{siunitx}
\usepackage{multirow}
\usepackage{booktabs}

\newtheorem{conjecture}{Conjecture}

\newcommand{\Red}  [1]{{\color{red}  {#1}}}

\newcommand{\x}{\mathbf{x}}

\newcommand{\pref}[1]{Eq.~(\ref{#1})}

\usepackage{amsmath,amsfonts,bm}

\def\eqref#1{equation~\ref{#1}}
\def\1{\bm{1}}

\DeclareMathAlphabet{\mathsfit}{\encodingdefault}{\sfdefault}{m}{sl}
\SetMathAlphabet{\mathsfit}{bold}{\encodingdefault}{\sfdefault}{bx}{n}

\newcommand{\E}{\mathbb{E}}

\newcommand{\R}{\mathbb{R}}

\newcommand{\kernel}{{NNPK}}

\newcommand{\nfk}{\text{\tiny NNPK}}
\newcommand{\linear}{{\sf \small Linear}}
\newcommand{\mlp}{{\sf \small MLP}}
\newcommand{\cnnsmall}{{\sf \small CNN-S}}
\newcommand{\cnnmed}{{\sf \small CNN-M}}
\newcommand{\lenet}{{\sf \small LeNet}}
\newcommand{\alexnet}{{\sf \small AlexNet}}
\newcommand{\vgg}{{\sf \small VGG19}}
\newcommand{\resnet}[1]{{\sf \small ResNet-#1}}
\usepackage{cleveref}
\begin{document}

\title{%1. Neural Random Features: Neural Networks as Kernels at Initialization\\
Learning from Randomly Initialized Neural Network Features
% 3. Learning from randomly initialized neural net output features\\
% 4. Logistic regression with random neural net features
}

% logistic regression with random nn features
% 




% \begin{icmlauthorlist}
% \icmlauthor{Ehsan Amid}{to}
% \icmlauthor{Rohan Anil}{to}
% \icmlauthor{Wojciech Kot\l{}owski}{goo}
% \icmlauthor{Manfred K. Warmuth}{to}
% % \icmlauthor{Fiuea Rrrr}{to}
% % \icmlauthor{Tateu H.~Yasehe}{ed,to,goo}
% % \icmlauthor{Aaoeu Iasoh}{goo}
% % \icmlauthor{Buiui Eueu}{ed}
% % \icmlauthor{Aeuia Zzzz}{ed}
% % \icmlauthor{Bieea C.~Yyyy}{to,goo}
% % \icmlauthor{Teoau Xxxx}{ed}
% % \icmlauthor{Eee Pppp}{ed}
% \end{icmlauthorlist}

\author{Ehsan Amid$^\dagger$\\
{\tt\normalsize eamid@google.com}
\and
Rohan Anil$^\dagger$\\
{\tt\normalsize rohananil@google.com}
\and
Wojciech Kot\l{}owski$^\star$\\
{\tt\normalsize kotlow@gmail.com}
\and
Manfred K. Warmuth$^\dagger$\\
{\tt\normalsize manfred@google.com}
}

\date{$^\dagger$Google Research, Brain Team\\
$^\star$Poznan University of Technology, Poznan, Poland}

% \icmlaffiliation{to}{Google Research, Brain Team}
% \icmlaffiliation{goo}{Poznan University of Technology, Poznan, Poland}

% \icmlcorrespondingauthor{Ehsan Amid and Rohan Anil}{\{eamid, rohananil\}@google.com}
% \icmlcorrespondingauthor{Eee Pppp}{ep@eden.co.uk}

% \icmlkeywords{Random Features, Neural Kernel, Kernel Methods}

% \vskip 0.3in
% ]




\maketitle 

\begin{abstract}
We present the surprising result that randomly initialized neural networks are good feature extractors in expectation. These random features correspond to finite-sample realizations of what we call Neural Network Prior Kernel (\kernel), which is inherently infinite-dimensional. We conduct ablations across multiple architectures of varying sizes as well as initializations and activation functions. Our analysis suggests that certain structures that manifest in a trained model are already present at initialization. Therefore, \kernel\ may provide further insight into why neural networks are so effective in learning such structures.
\end{abstract}
% \begin{abstract}
% B) For a given standard data set and neural net architecture
% (such as CFAR 10) we construct features from random
% initializations of the network by feeding the data set
% through the same network without training.
% The resulting feature vectors are then fed into a final softmax layer which is trained with a standard optimizer.

% Surprisingly we can show that by essentially just training the last layer, we can already get close to the fully trained performance of the neural net. Using visualization tools we also show that the same clusters already appear in the predictions based on random features as in the predictions of the fully trained network.

% Our method is robust across a large variety of standard data sets such as ... and modification of standard architectures.
% We also show that the performance improves with the number
% of random neural net features 
% and the same cannot be achieved with other feature constructions such as random sin/cos features.
% \end{abstract}
\section{Introduction: Neural Kernels}
There has been tremendous progress in recent years in connecting deep neural networks with kernel machines. The main example of such efforts is NTK~\citep{jacot2018neural} which introduces a kernel in terms of the gradient of the network with respect to the inputs. NTK has provided significant insight into the training dynamics~\citep{canziani2016analysis,novak2018sensitivity,li2018learning,du2019gradient,allen2019convergence} and generalization properties~\citep{arora2019fine,cao2019generalization} of neural networks. The properties of NTK are well studied in the infinite-width regime and under a squared loss assumption~\citep{lee2019wide,arora2019exact}. Recent progress has generalized these results to hinge loss and support vector machines~\citep{chen2021equivalence}. Also recently, it was shown that the training dynamics of any network trained with gradient descent induces a path kernel that depends on the NTK~\citep{domingos2020every}.

Another class of neural models involves constructing limit cases of infinitely wide Bayesian neural networks, resulting in a Gaussian process (NNGP)~\citep{lee2017deep,matthews2018gaussian,novak2018bayesian,garriga2018deep,borovykh2018gaussian}. NNGP describes the distribution of the output predictions of a randomly initialized infinitely-wide network. Adopting a sequential view of neural networks casts the uncertainty in the output of a layer as a conditional distribution on the previous layer's output. NNGP is closely related to NTK~\citep{jacot2018neural,hron2020infinite} and has been used to characterize the trainability of different architectures~\citep{schoenholz2016deep}.

Neural networks have been shown to have great expressive power even when the weights are randomly initialized ~\citep{frankle2018lottery,evci2020rigging,ramanujan2020s}. For instance, only training the BatchNorm~\citep{batchnorm} variables in a randomly initialized ResNet model provide a reasonable test performance on CIFAR and ImageNet-1k datasets~\citep{frankle2020training}. % 
Our work constructs random features from the outputs of repeatedly (randomly) re-initialized neural networks. We show that these features are in fact related to a kernel.
%Our work also focuses on the expressive power of randomly initialized neural networks through the lens of a new neural kernel.

\begin{figure}[t!]
\begin{center}
    \subfigure[]{\includegraphics[width=0.43\linewidth]{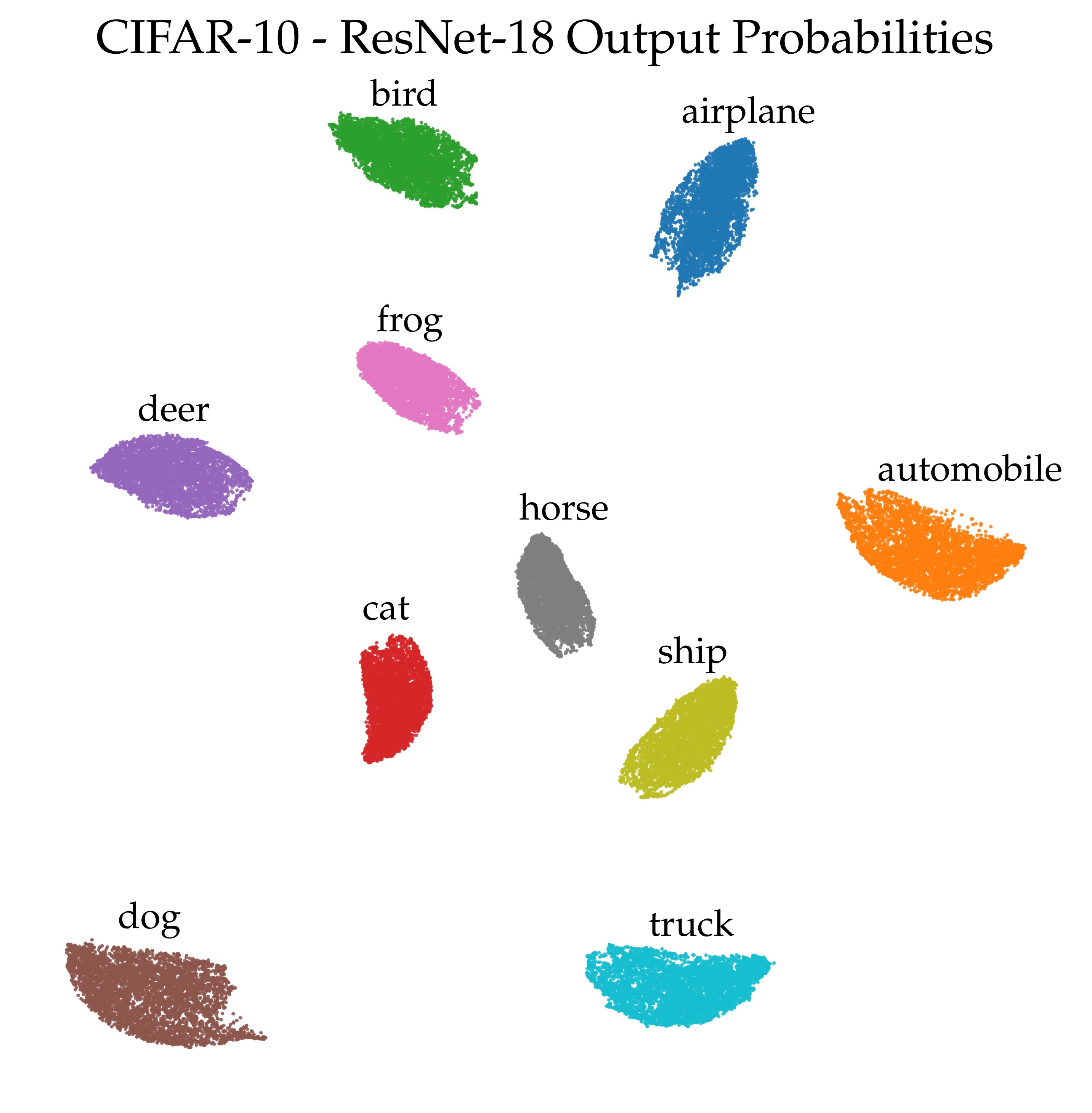}}
    \hspace{0.1cm} 
    \subfigure[]{\includegraphics[width=0.435\linewidth]{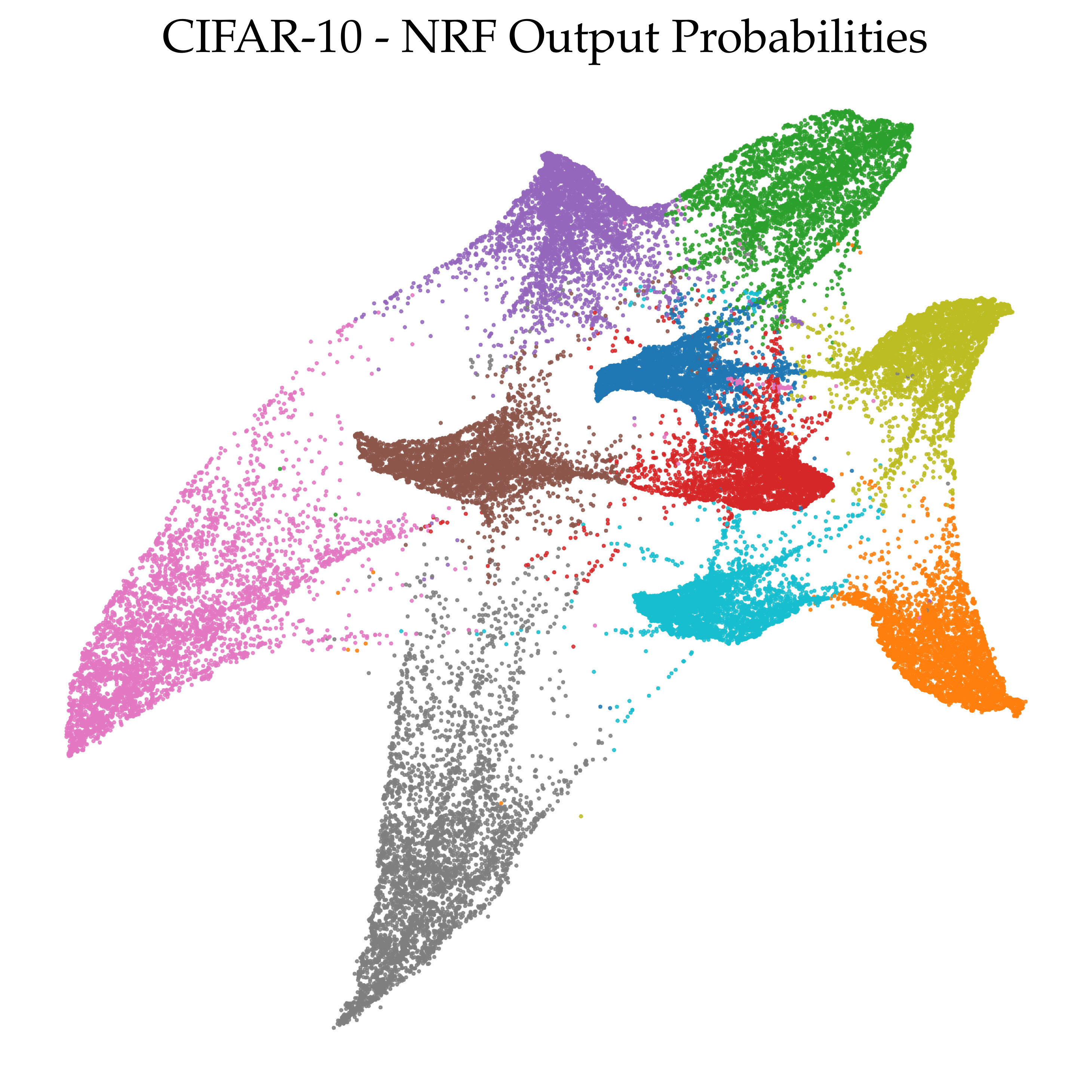}}
    \vspace{-0.25cm}
    \caption{TriMap~\citep{2019TRIMAP} visualization of the CIFAR-10 training examples: (a) output probabilities of a fully trained ResNet-18 model, (b) output probabilities of a (linear) logistic classifier trained on a \num{10240}-dimensional NRF representation, extracted from identical \emph{randomly initialized} ResNet-18 architectures.
    Note that the cluster in (a) already appear in (b) and the closeness of pairs of clusters is roughly preserved.}
    \label{fig:cifar_probs}
    \vspace{-0.55cm}
    \end{center}
\end{figure}
This paper introduces a new kernel construction induced from infinitely many randomly initialized neural networks. Our new kernel, called Neural Network Prior Kernel (\kernel), corresponds to the expected value of the inner-product of the logits of the network for two input examples. This expectation involves calculating an integral over all possible realizations of a prior distribution on the network's weights. Thus, \kernel\ inherently corresponds to an infinite-dimensional representation in a Hilbert space. Our finite-sample approximation of NNPK, called Neural Random Features (NRF), corresponds to embedding the input examples in the logits space using random networks, for which the weights are sampled from the prior (i.e., initial) distribution. Our extensive experiments with NRF across different architectures, initializations, activation functions, and dimensions suggest that NNPK is a useful tool to analyze the trainability of neural networks. Also, our experiments with NRF reveal that certain structures in the data that manifest in a fully trained neural network are already present at initialization. Thus, NNPK may also provide further insights into the expressive power of neural networks.

Figure~\ref{fig:cifar_probs} shows a TriMap~\citep{2019TRIMAP} visualization of the output probabilities of the training examples obtained from a ResNet-18 model, trained on the CIFAR-10 dataset along with the output probabilities of a linear classifier, trained on \num{10240} sampled features from \emph{randomly initialized} ResNet-18 models. The visualizations reveal a fair amount of similarity between the structure of the data and the placement of the clusters induced by the fully trained network and randomly initialized networks. In summary:
\begin{itemize}[noitemsep,topsep=0pt]
\item We formally define the Neural Network Prior Kernel (\kernel) and present its finite sample approximation as Neural Random Features (NRF).
\item We analyze NRF across a span of model architectures, initialization, and activation functions.
\item We report several intriguing observations that suggest that \kernel\ may reflect some properties about the trainability and generalization of the network at initialization before observing any data samples.
\item We also show that certain structures in the data that manifest in a fully-trained network are already observable at initialization using NRF. These findings hint that \kernel\ may provide further insights into understanding the expressivity of deep neural network architectures.
\end{itemize}

\section{Neural Prior Kernel}
We now formally define the Neural Network Prior Kernel (\kernel) of a neural network. Let $f^k_{\theta}: \R^d \rightarrow \R^k$ denote a neural network function parameterized by the weights $\theta \in \R^m$. The network transforms an input $x \in \R^d$ to the $k$-dimensional output $f^k_{\theta} \in \R^k$, called \emph{logits}. The logits are the output of a linear fully-connected layer and are fed into a softmax function to produce a $k$-dimensional probability vector over classes. The weights $\theta$ at the initialization follow a \emph{prior} distribution\footnote{The term \emph{prior} is commonly used in Bayesian statistics to describe the uncertainly of a distribution before observing some evidence. Here, we loosely use the term prior to denote the initial distribution of the model weights, although the connection to Bayesian statistics is yet to be established in future work.} $\pi(\theta)$. For deep neural network, this distribution usually amounts to a scaled (truncated) normal or a uniform distribution. The \kernel\ of the network between two inputs $x, x'\in \R^d$ is then defined in terms of an expectation with respect to the initial weight distribution $\pi$,
\begin{equation}
    \label{eq:NFK}
    % \boxed{
    \begin{split}
    \kappa_{\nfk}(x, x') & = \E_{\theta\sim \pi}[\langle\,f^k_{\theta}(x),f^k_{\theta}(x')\rangle]\, .\\
    \end{split}\tag{\kernel}
    % }
\end{equation}
That is, \kernel\ is simply the expected value of the inner-product of the output logits for the given two inputs. Clearly, \kernel\ is a valid kernel since it corresponds to a convex combination of symmetric and positive semi-definitive functions. \kernel\ is also deterministic for a given distribution $\pi$, as the expectation is over all realizations of $\theta \sim \pi$.

\begin{figure*}[t!]
\begin{center}
    \includegraphics[height=0.46\linewidth]{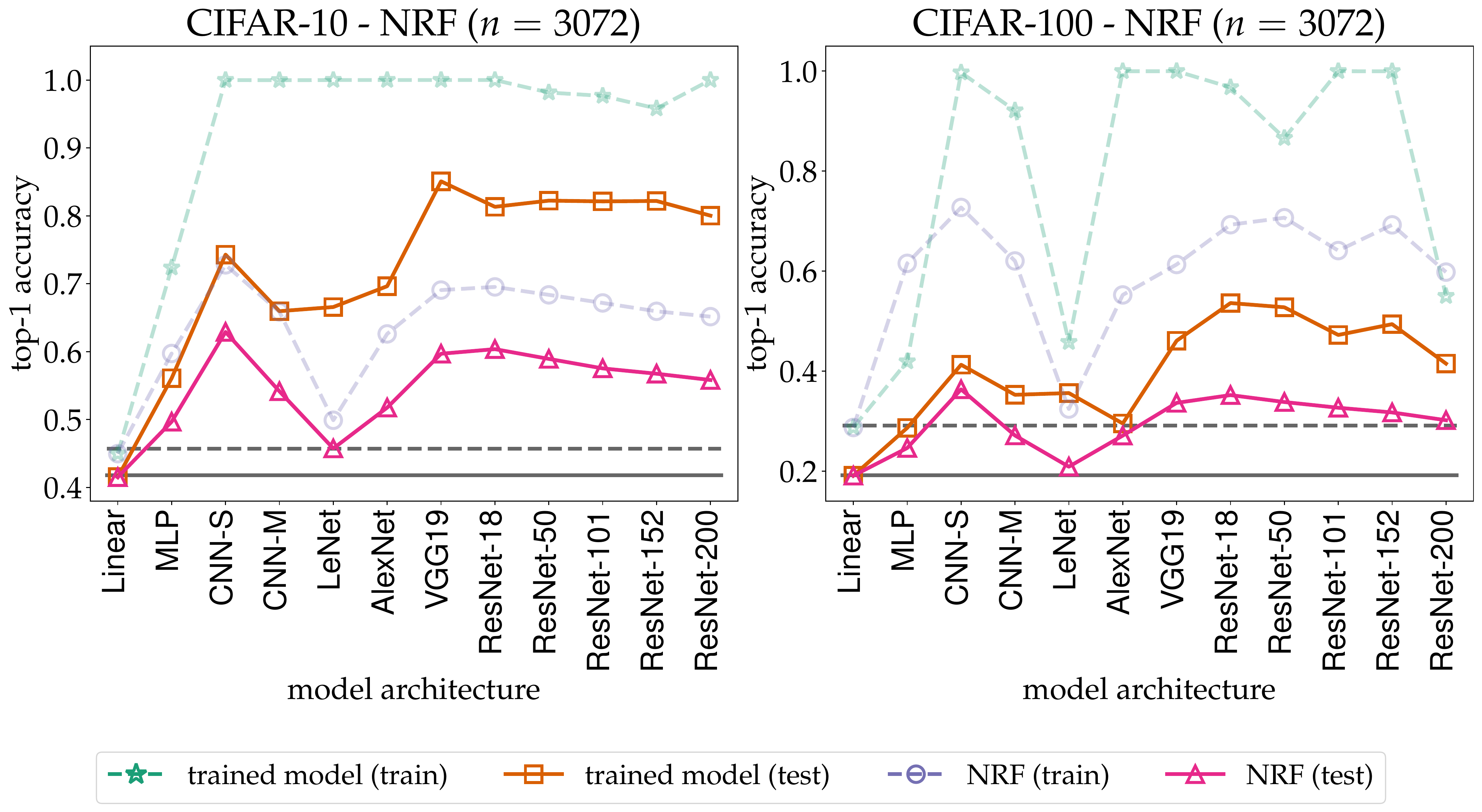}
    % \hspace{0.1cm} 
    % \subfigure[]{\includegraphics[width=0.47\linewidth]{figs/cifar100_architectures.pdf}}
    \vspace{-0.3cm}
    \caption{NRF of different model architectures on CIFAR-10/100 datasets: Train and test accuracy of linear classifiers trained on randomly sampled $n=3072$ NRF using different standard architectures. The accuracy of the trained networks (without data augmentation) is also shown for each network. The models are sorted roughly in the order of their capacity. We also plot the train and test accuracy of a linear classifier on the original CIFAR-10/100 inputs using dashed and solid horizontal lines, respectively.
    Note that the train and test curves for learning with NRF present very rough approximations of the corresponding curves for the fully trained models.}
    \label{fig:arch}
    \vspace{-0.45cm}
    \end{center}
\end{figure*}

We can calculate a finite sample approximation of \kernel\ as
\begin{equation}
    \label{eq:NFK-sample}
    \begin{split}
    \hat{\kappa}_{\nfk}(x, x') & = \frac{1}{n}\, \sum_i\langle f^k_{\theta_i}(x), f^k_{\theta_i}(x')\rangle\, ,
    \end{split}
\end{equation}
using $n$ iid samples $\{\theta_i\}_{i=1}^n$ drawn from $\pi$. For a given neural network architecture, this corresponds to randomly re-initializing the networks $n$ times and calculating the average inner-product between the logits for the given two inputs. 

For a randomly initialized network, the output logits of an example are correlated. In order to obtain uncorrelated output samples, in our construction, we set $k = 1$ while keeping the rest of network architecture unchanged. In this case, each randomly initialized network maps each input example to a single scalar value. Thus, \pref{eq:NFK-sample} can be written as
\begin{equation}
    \label{eq:NFK-phi}
    \begin{split}
    \hat{\kappa}_{\nfk}(x, x') & = \langle \phi_n(x), \phi_n(x')\rangle\, ,
    \end{split}
\end{equation}
where the embedding map $\phi_n: \R^d \rightarrow \R^n$ is defined as
\begin{equation}
    % \label{eq:phi}
    \phi_n(x) = \sfrac{1}{\sqrt{n}}\, [f^1_{\theta_1}(x), \ldots, f^1_{\theta_n}(x)]^\top\, .\tag{NRF}
\end{equation}
These randomly generated features are then treated as new representations for the training examples and are used to train a classifier. In our experiments, we mainly focus on training a linear classifier with a softmax output to predict the class probabilities. At inference, the same randomly initialized networks are used to construct the random features for a test example.

In the following sections, we show that \kernel\ is a non-trivial kernel which allows separating the data significantly better than the original input space. We analyze the dependency of \kernel\ (via its finite-sample approximation NRF) on different elements and properties of the models. Interestingly, \kernel\ is a fixed kernel for a given architecture and therefore, the same randomly initialized networks to realize NRF can be used across different datasets. This is in contrast to the \emph{data-dependent} path kernel~\citep{domingos2020every} which relies on the training trajectory of the model.

\section{Experiments}
We conduct several ablations to understand the effect of different elements, such as architecture, width, depth, activation function, etc., on the \kernel. For the first set of experiments, we conduct ablations on CIFAR-10/100 datasets of images~\citep{cifar100}. To show the generality of the NRF extracted from the \kernel, we also perform an experiment on the ImageNet-1K dataset~\citep{imagenet}.

\subsection{Effect of Architecture}
We first show the dependency of NRF on different model architectures. For this, we consider a number of different models of varying sizes.
These models include a two-layer fully-connected network (\mlp), a small (\cnnsmall) and a medium (\cnnmed) convolutional network, variants of \lenet~\citep{lecun1989backpropagation}, \alexnet~\citep{krizhevsky2012imagenet}, and \vgg~\citep{simonyan2014very} as well as several ResNets~\citep{resnet} (\resnet{18}, \resnet{50}, \resnet{101}, \resnet{152}, and \resnet{200}). The details of all models are described in the appendix. We also consider a linear projection of the input data using a random matrix. The \linear\ baseline corresponds to a random projection of the input data which is used for applications such k-neareast neighbor search~\citep{kleinberg1997two} and random projection trees~\citep{dasgupta2008random}. We use a He normal initialization~\citep{he2015delving} for the ResNet models and use Glorot normal~\citep{glorot2010understanding} for the rest.

We set the NRF dimension $n$ equal to the original dimensionality of the input data, which is $32 \times 32 \times 3 = 3072$. This way, any improvement in separability over the original input data is solely due to a better representation by NRF. To classify the data, we train a logistic regression classifier on NRF using the L-BFGS~\citep{liu1989limited} optimizer for which we tune the $\mathrm{L}_2$-regularizer value. We repeat each experiment over 5 random trials.

\begin{figure*}[t!]
\begin{center}
\subfigure[]{\includegraphics[width=0.4\linewidth]{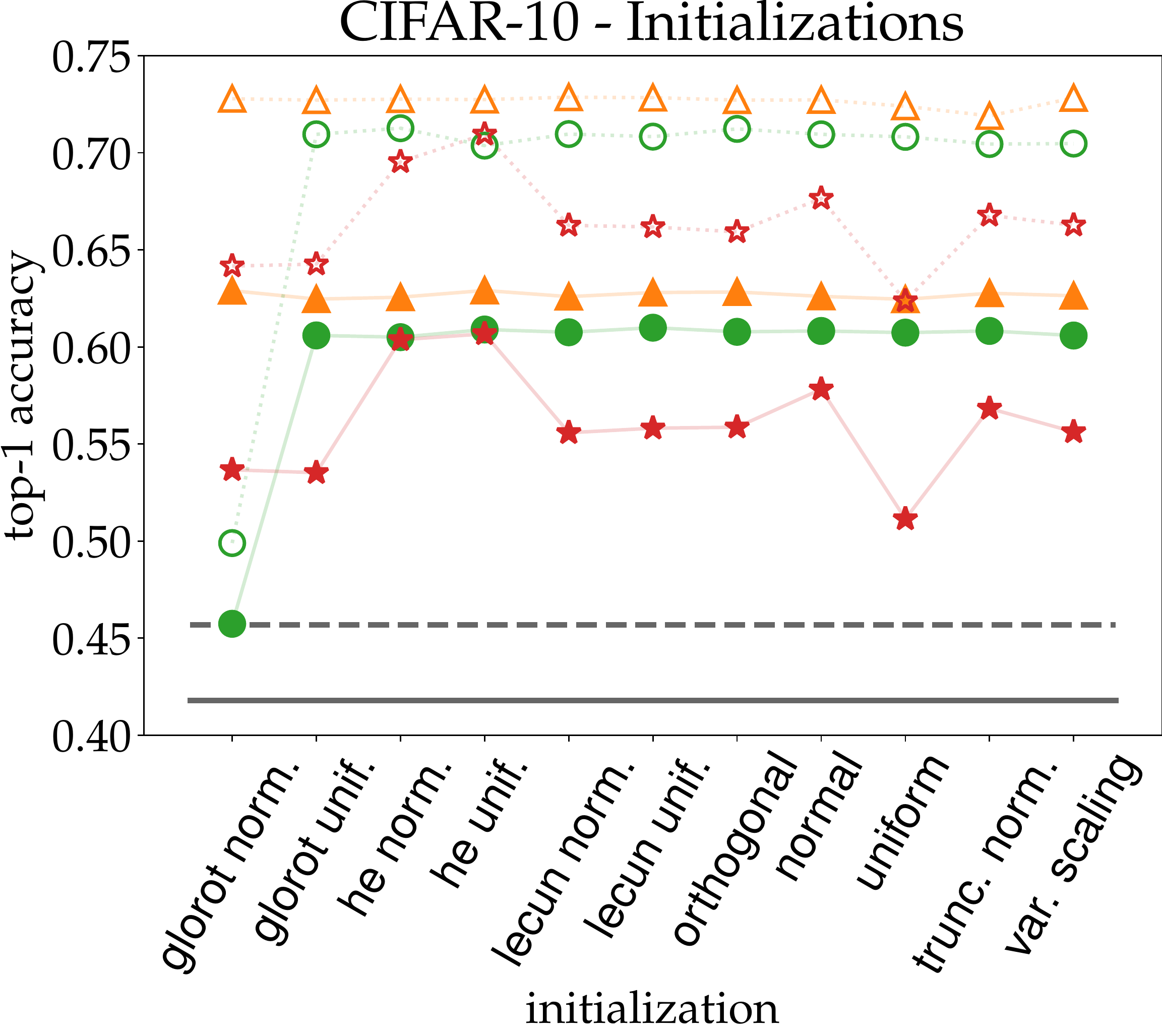}}
    \;\;\;\;\;\;\subfigure[]{\includegraphics[width=0.4\linewidth]{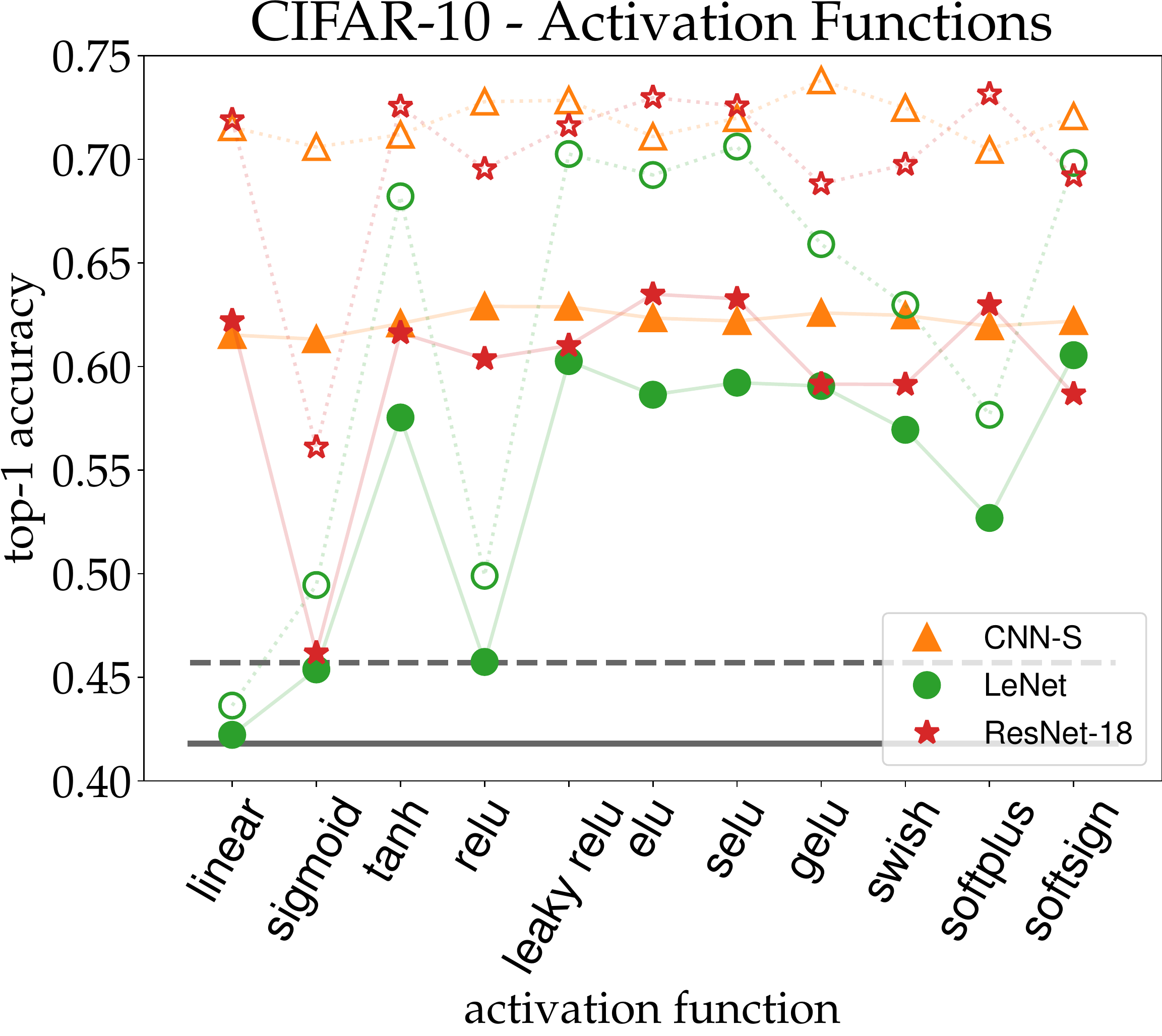}}
    % \hspace{0.1cm} 
    \vspace{-0.4cm}
    \caption{Effect of initialization and activation function on NRF on CIFAR-10: Train (unfilled) and test (filled markers) accuracy of linear classifiers on $n=3072$ NRF sampled from \cnnsmall, \lenet, and \resnet{18} architectures using different (a) initializations and (b) activation functions. The train and test accuracy of a linear classifier on the original CIFAR-10 inputs are shown using dashed and solid horizontal lines, respectively.
    Note that for fully trained ResNet networks, He normal initialization and ELU activations are known to achieve good performance~\citep{he2015delving,shah2016deep}. Surprisingly our NRF methods picks out these choices in (a) and (b).}
    \label{fig:inits_acts}
    \vspace{-0.5cm}
    \end{center}
\end{figure*}

In Figure~\ref{fig:arch}, we plot the train and test accuracy of NRF along with the baseline linear classifier trained on the original input data (dashed and solid horizontal lines, respectively). We also plot the train and test accuracy of the trained network. To train each model on the training examples, we use a SGD with Nesterov momentum optimizer~\citep{nesterov} with a batch size of 128 for 100k iterations. We use a linear warmup followed by a linear decay schedule for the learning rate and tune the maximum learning rate value, momentum constant, and the weight decay parameter for 128 trials using a Bayesian parameter search package. One important detail to notice is that we train the networks~\emph{without data augmentation}. Data augmentation is a standard practice for training such networks and is a strong form of regularization that allows larger networks provide better generalization when the number of train examples is limited. With data augmentation, we expect the performance of the models to increase with size, which is evident in several existing benchmarking efforts on CIFAR-10/100 datasets.\footnote{For instance, see~\url{https://github.com/kuangliu/pytorch-cifar} and~\url{https://paperswithcode.com/sota/image-classification-on-cifar-10} for some benchmarking results.} However, our goal here is to understand how much a trained network can improve on the initial representation of the data, reflected in \kernel, \emph{after} seeing the (original) training examples.

Several interesting observations can be made from Figure~\ref{fig:arch}. First, the NRF extracted from all architectures perform better than the baseline classifier trained on the original input image representations. The performance of \linear\ random projection is about the same as the original data, as we expect such projections to be at best as good as the original data~\citep{dasgupta2003elementary}. Next, the performance of NRF is not monotonic with the architecture size. Among the models, the shallow \cnnsmall\ model provides the best performance while performance drops by expanding the model to \cnnmed\ or \lenet. A similar phenomenon happens when we increase the ResNet sizes. More interestingly, the performance of the trained model also follows a similar pattern to NRF on both datasets. This suggests that a trained network (without data augmentation) improves on the initial representation of the input that is reflected via \kernel\ at initialization. Thus, the performance after observing the inputs is proportional to the initial \kernel\ representation. We test this hypothesis further in our later experiments.

\subsection{Effect of Initialization}
\label{sub:init}
The expectation in the definition of \kernel\ is with respect to the prior distribution (i.e., initialization) of the model weights. In order to analyze the effect of the prior distribution $\pi$, we evaluate the NRF for \cnnsmall, \lenet, and \resnet{18} models using different standard weight initializations. As before, we set the embedding dimension $n=3072$. Figure~\ref{fig:inits_acts}(a) shows the train (unfilled) and test (filled markers) accuracy of the linear classifiers trained on NRF. Similarly, we show the baseline train and test accuracy (dashed and solid lines, respectively) on the original inputs.

From~\ref{fig:inits_acts}(a), we observe that the NRF of some networks, e.g. \cnnsmall\, are more robust to the choice of initialization than the others. The NRF of \resnet{18} architecture, in particular, seem to vary significantly and are degraded, especially when using (Glorot) uniform or Glorot normal initializations. In addition, the NRF induced by the \resnet{18} model seems to perform the best when using He normal (which is usually the default initialization for ResNets~\citep{he2015delving}).
\begin{figure*}[t!]
\begin{center}
    \includegraphics[height=0.43\linewidth]{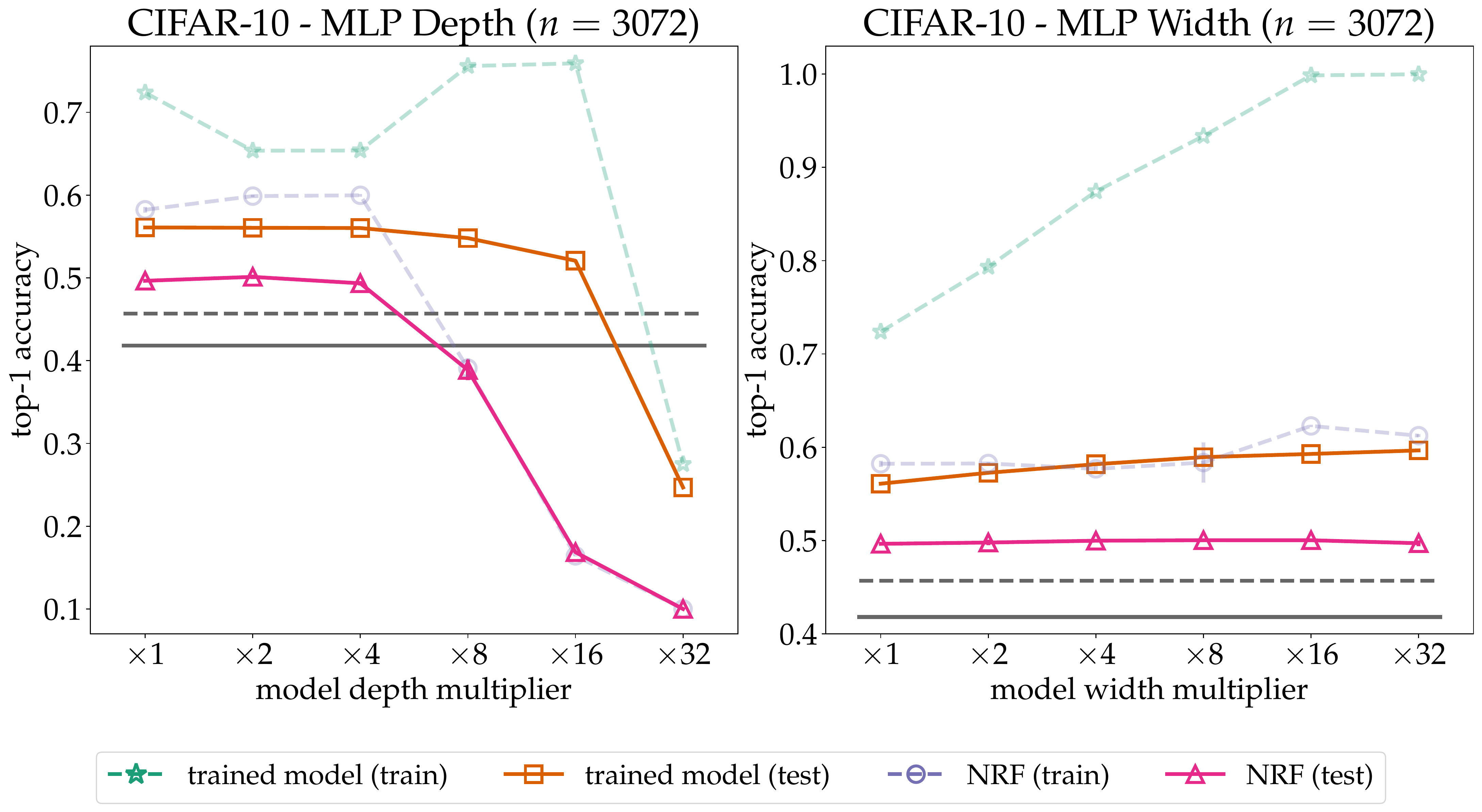}\\
    \hspace{1cm}(a) \hspace{6.5cm} (b)
    % \hspace{0.1cm} 
    % \subfigure[]{\includegraphics[width=0.47\linewidth]{figs/cifar100_architectures.pdf}}
    \vspace{-0.25cm}
    \caption{Effect of depth and width on NRF on CIFAR-10: Train and test accuracy of linear classifiers on $n=3072$ NRF extracted from an MLP model with varying (a) depth and (b) width. We also plot the train and test accuracy of the trained networks (without data augmentation). We also show the train and test accuracy of a linear classifier on the original CIFAR-10 inputs using dashed and solid horizontal lines, respectively.
    Note again that the NRF curves are rough approximations of the curves for the fully trained models.}
    \label{fig:depth_width}
    \vspace{-0.5cm}
    \end{center}
\end{figure*}
\subsection{Effect of Activation Function}
Apart from the prior distribution on the weights (i.e., initialization), NRF is also dependant on the activation functions of the network. To study their effect, we conduct an experiment by varying the activation functions of \cnnsmall, \lenet, and \resnet{18} models. Again, we set the embedding dimension $n=3072$ and use the default initializations for each model.

From Figure~\ref{fig:inits_acts}(b), we can see that certain activation functions such as leaky ReLU ($\text{slope}=0.1$) seem to consistently produce better NRF across architectures (although ELU seems to perform the best on \resnet{18}, which is known to also perform well for a trained ResNet~\citep{shah2016deep}). On the other hand, the NRF induced by the sigmoid activation function is consistently worse for all architectures. This suggests that the choice of a good activation function has a significant effect on the initial representation of the model, thus it may as well affect the generalization performance of the model after training.

In summary, based on our experiments on different initializations and activation functions and our findings on \resnet{18}, \kernel\ seems to hint at the choices that are known to work also well for fully trained models~\citep{DecouplingBU,shah2016deep,martens2021rapid}.
% I think you really want to hammer home the correlation (in the Activation Function, Effect of Depth and Width, and Effect of Dimension) between NRF performance and fully-trained model performance. Consider adding an extra sentence to each of these ablations really driving this point home -- it seems like your conclusion/discussion relies on it.

\subsection{Effect of Depth and Width}
A natural question is whether depth or width affects the \kernel\ of a network. To study this, we create different variants of our basic \mlp\ model by expanding its width or depth (i.e. number of layers) by different multiples. Figure~\ref{fig:depth_width} illustrates the train and test accuracy of these expanded models as a function of the multiplier. We use the default Glorot normal initialization for the networks. We also plot the train and test accuracy of the trained models (using the same tuning protocol and without data augmentation). Interestingly, the \kernel\ does not improve by increasing the depth of the network. Also, the deeper networks become harder to train without data augmentation and other regularization techniques. In contrast, both the performance of \kernel\ and the trained network improves slightly by increasing the width. This is also consistent with the empirical findings on wide neural networks~\citep{novak2018sensitivity,canziani2016analysis}.

\subsection{Effect of Dimension}
\begin{figure*}[t!]
\begin{center}
    \includegraphics[height=0.48\linewidth]{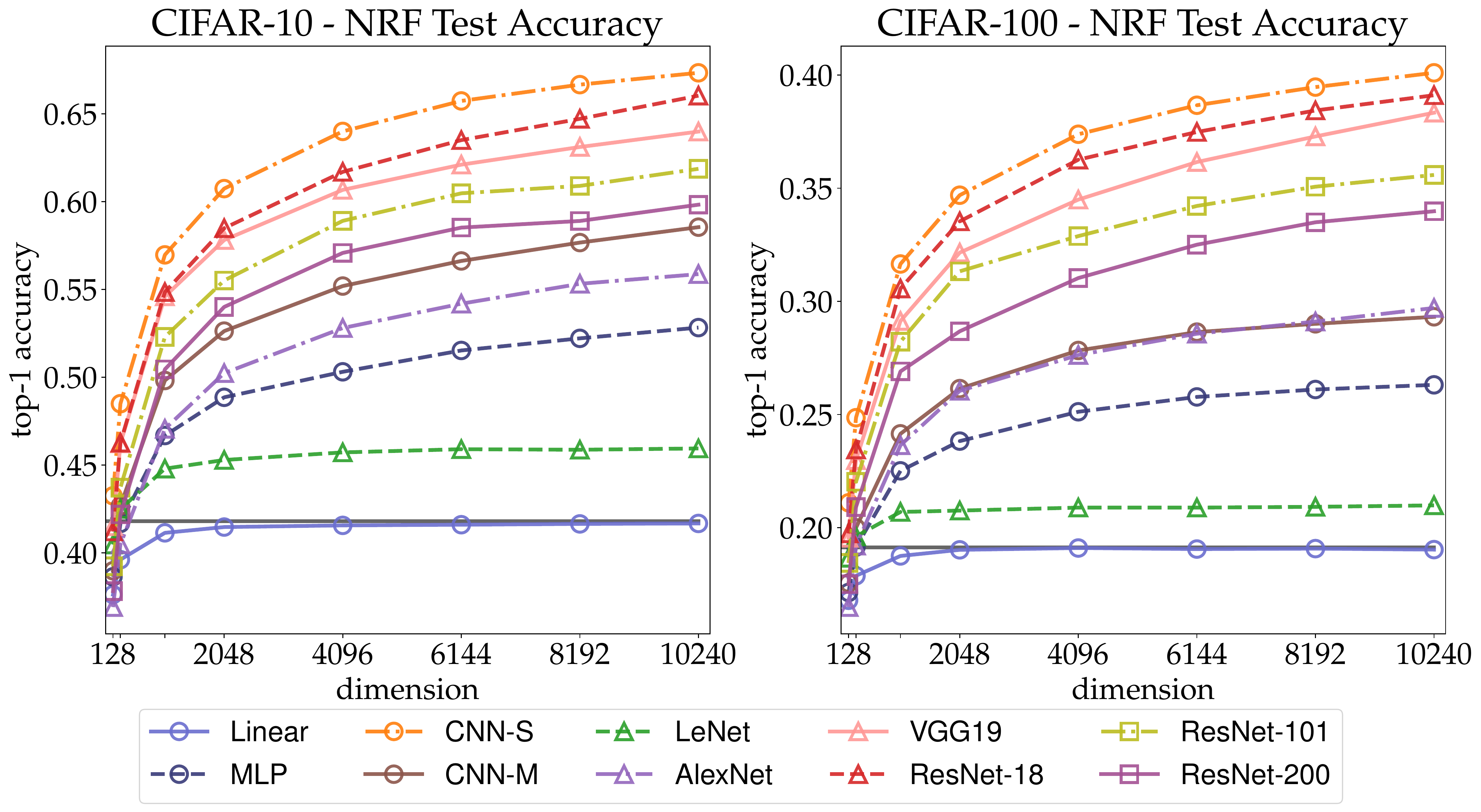}\\
    \hspace{1.5cm}(a) \hspace{6.5cm} (b)
    % \hspace{-0.2cm} 
    % \subfigure[]{\includegraphics[height=0.455\linewidth]{figs/cifar10_test_dims.pdf}}
    \vspace{-0.25cm}
    \caption{Effect of dimension $n$ on the finite-sample approximation of \kernel\ using NRF: Test accuracy of linear classifiers on (a) CIFAR-10 and (b) CIFAR-100, trained on NRF with varying dimensionality $n$ extracted from different model architectures. We also plot the test accuracy of the linear model trained on the original CIFAR-10/100 input images (solid horizontal lines). As we expect, the accuracy of the linear random projection of the input images (\linear) does not exceed the accuracy on original inputs, as we increase the dimension. On the contrary, the accuracy of the classifiers trained on NRF improves monotonically with dimension. This justifies the definition of the \kernel\ as a limit case when $n\rightarrow \infty$. Also, the order of the performance using different model architectures remains roughly the same across dimension.}
    \label{fig:cifar_dim}
    \vspace{-0.5cm}
    \end{center}
\end{figure*}

% "We show that" -> :'(. This is such a big claim. You can do "For our analysis, we found...", but we show w/o qualifiers implies universality.
We empirically show that the finite-sample approximation of \kernel\ using NRF in~\pref{eq:NFK-sample} improves for all networks as we increase the number of samples $n$. This corresponds to increasing the embedding dimension of NRF in $\phi_n$. Figure~\ref{fig:cifar_dim} shows the test accuracy of linear classifiers trained on NRF using different dimensions $n$ on CIFAR-10/100 datasets. As we expect, the performance of the random projection of the data (\linear) matches the performance of the linear classifier trained on the original input features (solid horizontal lines) and does not improve with dimension. On the contrary, the performance of NRF increases monotonically with dimension for all networks. This justifies the definition of the \kernel\ as a limit case when $n\rightarrow \infty$. Interestingly, the order of performance for different models (along the vertical axis) remains roughly the same across different dimensions. This suggest that the observations about \kernel\ are relatively consistent using different dimensionality $n$ to approximate the kernel.

\subsection{Does Skip Connection Improve NRF?}
The immense success of ResNets is mainly attributed to two elements: the use of Batch Normalization (BatchNorm)~\citep{batchnorm} and skip connections~\citep{resnet}. There has been a number of efforts to replace or remove these two elements from ResNets~\citep{zhang2019fixup,gaur2020training,bachlechner2021rezero}. The more recent adjustments proposed in~\citep{martens2021rapid} have shown further promise. This modification includes replacing the ReLU activation function with a scaled leaky ReLU and the use of delta orthogonal initialization for the convolutional kernels (and orthogonal initialization for the final dense layer). Here, we explore the dependence of \kernel\ and the trained models (without data augmentation) on skip connections (and BatchNorm). One important point to notice is that BatchNorm is an affine transformation on the activations in each layer during inference. Since the  mean and scale values are initially set to $0$ and $1$, respectively, BatchNorm with default initialization will have no effect on the \kernel, but affects the performance of the trained model. We also examine whether NRF varies by removing skip connections and whether the changes proposed in~\citep{martens2021rapid} affect the results.

\begin{figure*}[t!]
\begin{center}
    \subfigure{\includegraphics[width=0.60\linewidth]{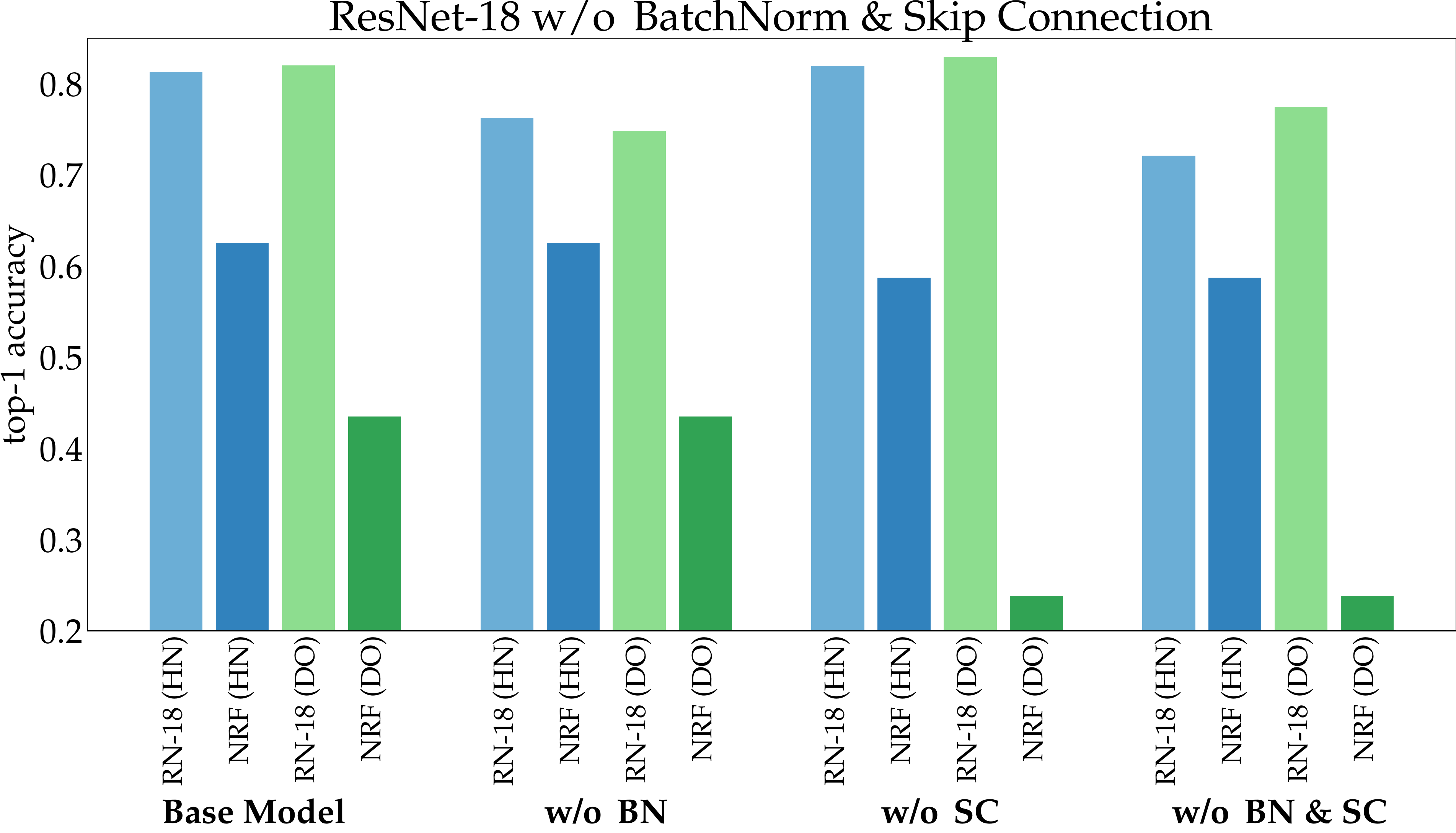}}
    % \hspace{0.1cm} 
    % \subfigure[]{\includegraphics[width=0.48\linewidth]{figs/cifar10_bn_skip.pdf}}
    \vspace{-0.2cm}
    \caption{Effect of BatchNorm and skip connection on the \resnet{18} model and the induced NRF on CIFAR-10: we consider the original \resnet{18} model with He normal initialization and ReLU activations (RN-18 (HN)) along with the variant proposed in~\citep{martens2021rapid} with delta orthogonal initialization and scaled leaky ReLU activations (RN-18 (DO). We also generate $n=3072$ NRF using each model. We plot the test accuracy of each trained model (without data augmentation) as well as the performance of linear classifiers trained on NRF for the base models along with the variants in which BatchNorm and/or skip connections are removed. As we expect, we observe that BatchNorm with the default initial values behaves as an identity map, thus has no effect on NRF. On the other hand, removing skip connections degrades the performance on NRF.}
    \label{fig:bn_skip}
    \vspace{-0.5cm}
    \end{center}
\end{figure*}

We create variants of the \resnet{18} model by removing BatchNorm and/or skip connections. In the first approach, we use the default He normal initialization with ReLU activation while in the second case, we use the delta orthogonal initialization proposed in~\citep{martens2021rapid} for the convolutional filters (and orthogonal initialization for the final dense layer). We also replace the ReLU activation with the scaled leaky ReLU proposed in~\citep{martens2021rapid}. We set the slope of the leaky ReLU to $0.3$. For both variants, we train the models without data augmentation and tune the learning rate, momentum constant, and the weight decay parameters for $128$ trials. We generate $n=3072$ dimensional NRF using each variant and train linear classifiers. The results are shown in Figure~\ref{fig:bn_skip}.

It can be seen from Figure~\ref{fig:bn_skip} that the performance of the trained base model (using He normal initialization and ReLU activation) degrades without BatchNorm (and skip connection). However, the performance remains about the same without only the skip connections. As we expect, the performance using the NRF of the base model does not vary by adding/removing BatchNorm. However, removing skip connections degrades the performance of NRF. 

In comparison, the trained model using delta orthogonal initialization and leaky ReLU activation performs better than the base model in most cases, especially when both BatchNorm and skip connections are removed. The NRF of this model also follows a similar pattern: the NRF does not vary by adding/removing BatchNorm, but degrades without the skip connections. However, the performance of the NRF of this model is comparatively lower than the NRF of the base model. 

\begin{figure*}[t!]
\begin{center}
    \vspace{.7cm}
    % \subfigure[]{\includegraphics[height=0.2\linewidth]{figs/cifar10_bn_skip.pdf}}
    % \hspace{0.1cm} 
    \subfigure{\includegraphics[height=0.5\linewidth]{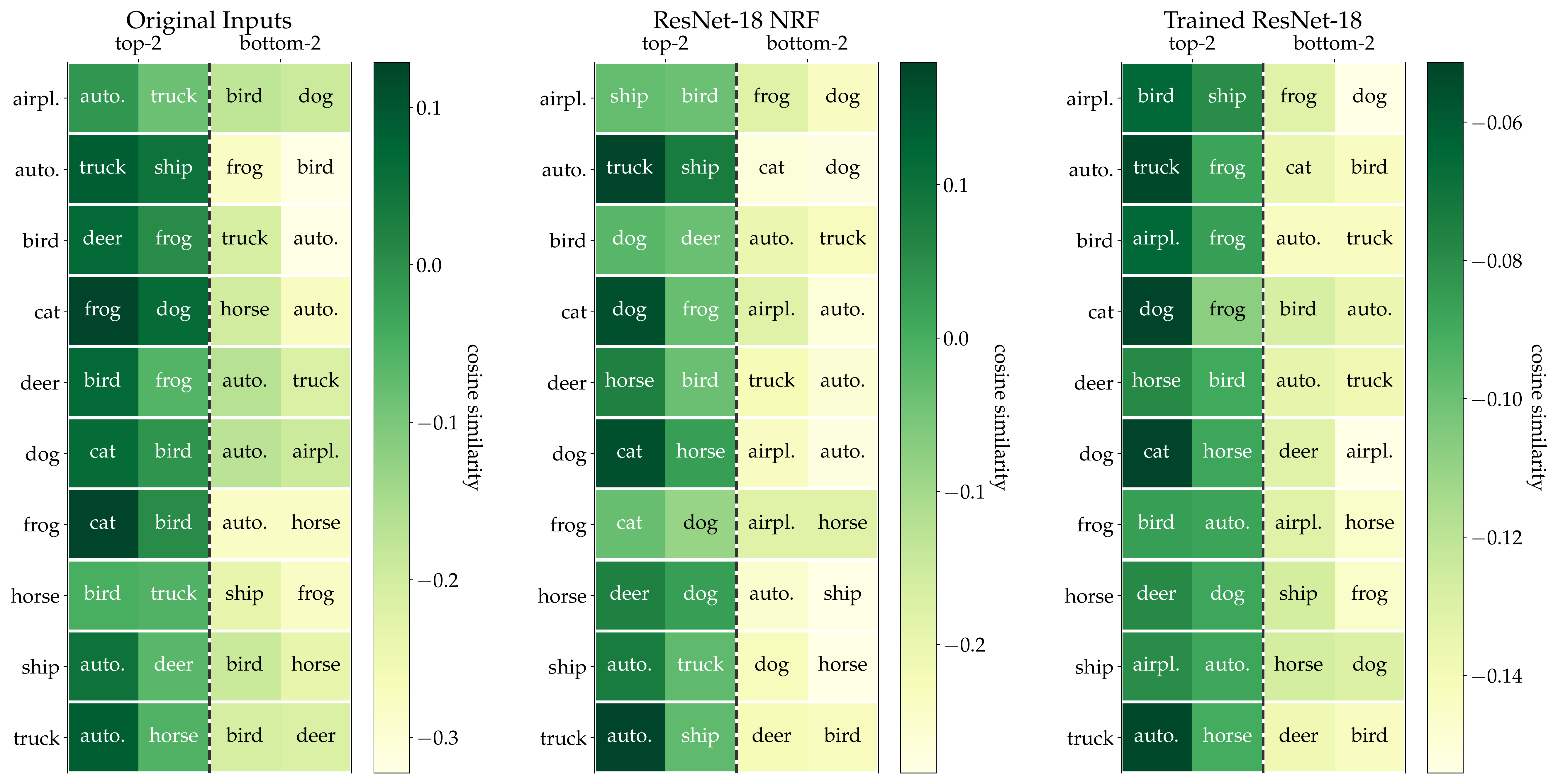}}
    \vspace{-0.25cm}
    \caption{Cosine similarities for different classes of ResNet-18: a linear classifier trained on the original inputs (left), on NRF sampled from ResNet-18 (middle), and the softmax layer of a fully trained ResNet-18 network (right). NRF appear to be an intermediate representation between the original input data and the representation induced by a fully trained model. Some classes that are close in the fully trained model are closer in the NRF representation compared to the original input representation.}
    \label{fig:cosine}
    \vspace{-0.5cm}
    \end{center}
\end{figure*}

These observations are interesting from several aspects. First, the NRF seems to improve by using skip connections for both initialization and activation function combinations. However, the performance of the trained model using the adjustments proposed in~\citep{martens2021rapid} seems to improve without the skip connections. This observation shows that there are cases for which \kernel\ may not reflect the performance of the final trained model (although the results may vary when using data augmentation and other types of regularizations). Additionally, as we expect, the \kernel\ is not affected by BatchNorm using the default initialization for the BatchNorm variables. However, the dependence of \kernel\ on different types of normalization techniques such as Layer Normalization~\citep{ba2016layer} and Instance Normalization~\citep{ulyanov2016instance} is yet to be examined.

\subsection{\kernel\ Reveals Structure}
In this section, we show that the structure in the data that appears in a fully-trained network is somehow reflected in \kernel. In other words, \kernel\ seems to be an intermediate representation between the original input space and the final trained network. For this, we consider the linear classifier trained on the original input images from the CIFAR-10 dataset. We also consider the fully trained \resnet{18} model (without data augmentation) as well as the linear classifier trained on NRF, sampled from the randomly initialized \resnet{18} models. We set $n=3072$ the same as the original data dimension. We then calculate the cosine similarities of the class weights for each classifier (for the fully trained \resnet{18}, this corresponds to the weights of the output softmax layer). We then plot the top and bottom-2 similar classes for each of the 10 classes of CIFAR-10 in Figure~\ref{fig:cosine}.

We can see from Figure~\ref{fig:cosine} that in many cases (such as the class `horse'), the top similar classes are identical for the NRF classifier and the output layer of the fully trained network. However, in some cases (e.g., `automobile'), the top classes of the classifier trained on the original CIFAR-10 images are closer to the NRF classifier than the \resnet{18} model. This observation further hints at the following conjecture.
% This is a not-insignificant leap. You essentially go from saying "Figure 7 looks promising! But actually not so much for other cases before throwing out a conjecture which ignores the "not so promising" cases (i.e. cases where softmax acting on original inputs are more similar to Trained ResNet-18 than the NRF version). If you want to make this claim, perhaps remind the reader of all the ablation studies where NRF performance was highly correlated with model performance, rather than relying on the weak evidence presented in Figure 7?
\begin{conjecture}
The original representation of the data is first projected via \kernel\ into an initial representation which is then enhanced throughout the training into the final representation of the fully-trained network.
% "into an intermediate representation" => into an initial representation? Kind of like it's not intermediate, but rather an initial structuring of the data that's successively refined via gradient descent?
\end{conjecture}
A further theoretical and empirical study of the above conjecture is an interesting future research direction.

% \subsection{Augmentation?}
\subsection{Results on the Larger ImageNet-1K Dataset}
In order to demonstrate the applicability of the results using NRF on a larger scale, we perform an experiment on the ImageNet-1K dataset~\citep{imagenet}. The dataset contains around 1.2M training images from \num{1000} classes. A linear classifier on the original input images (150K features) achieves a 3.4\% test top-1 accuracy. In contrast, a linear classifier trained on \num{4096} NRF, sampled from randomly initialized ResNet-18 models, achieves a 10.3\% test accuracy. This result is fascinating given that NRF achieves a significantly higher accuracy with 37$\times$ fewer features than the original input features. Training a linear classifier and a two-layer MLP on \num{31568} NRF achieves 12.2\% and  15.2\%  top-1 test accuracies, respectively.

\section{Conclusion}
% Consider changing: "we find from experiments..." => Our empirical analysis across a span of architectures, activation functions, and other ablations suggests the structures constructed throughout training by neural networks are initiated from a quintessential random kernel (or something like this). 
From experiments on a large variety of image classification datasets and neural network architectures, we found that the representation of a fully trained network is already ``echoed'' in a representation induced from random neural network features. These random features are constructed from multiple randomly initialized networks, and a simple linear classifier is trained on top of these features for classifying the input.

% Consider changing to: "This line of work prompts several natural questions:" 
% Also consider bolding the questions in \item -- this'll help distinguish the questions you're asking from the answers you're providing.
This line of work leads to several empirical questions: 
\begin{itemize}[noitemsep,topsep=0pt]
    \item Can the NRF be used to find suitable architectures for a given dataset rapidly? Figures \ref{fig:arch}-\ref{fig:cifar_dim} suggest that this is possible; in each case, the comparative performance on the fully trained network is reflected by NRF.
    % \item What are the key architectural features that make the NRF reflect the fully trained network performance?
    \item Which properties affect (thus, can be measured by) \kernel? What are the key architectural features that make the NRF reflect the performance of the fully trained network?
    % I think you already hypothesize an answer to this question in your conjecture? 
    % \item Can NRF be used as a sub-module to find smaller networks with competitive overall prediction performance? 
    % My two cents: stick to 3 really strong questions and axe the weakest one of these. 
    \item Are there other feature extraction methods that can achieve the same echoing of the performance of the fully trained model?
    % I think "are there other *shortcuts* which achieve similar insight into fully trained network performance" may be interesting (since then you're tying this into the field of NAS). 
\end{itemize}
\bibliography{refs}
\bibliographystyle{icml2020}

\newpage
% \onecolumn
% \title{Learning from Randomly Initialized Neural Network Features\\(Supplementary Material)}
\appendix

\section{Details of the Model Architectures}

\paragraph{\mlp\ Model:} contains two fully-connected layers of size $128$ with ReLU activations followed by the output layer.

\cnnsmall\ \textbf{model:} contains two 2-D convolutional layers of size 32 and 64, respectively, with $5\times 5$ filters. Each convolutional layer is followed by a max-pooling layer. We then apply a dense layer of size 512 followed by the final dense layer. All the activation functions are set to ReLU.

\cnnmed\ \textbf{model:} contains four $5\times 5$ convolutional layers, for which the sizes are 32, 64, 64, and 32, respectively. The first two convolutions are each followed by a max-pooling layer. We then similarly apply a dense layer of size 512 followed by the final dense layer. All the activation functions are set to ReLU.

\lenet\ \textbf{model:} is a variation of the \lenet\ model described in~\url{https://www.kaggle.com/blurredmachine/lenet-architecture-a-complete-guide} in which we replace the activation functions with ReLU.

\alexnet\ \textbf{model:} is a variation of the model desctibed in~\url{https://www.analyticsvidhya.com/blog/2021/03/introduction-to-the-architecture-of-alexnet/} where we changed the filter sizes in the first and last convolutional layers to $3\times 3$ and $1\times 1$, respectively. 

\vgg\ \textbf{model:} we use the VGG-19 implementation available in Keras at~\url{https://www.tensorflow.org/api_docs/python/tf/keras/applications/vgg19/VGG19}.

{\sf \small ResNet}\ \textbf{models:} are standard ResNet-v1 models implemented in TensorFlow.
\end{document}